\documentclass{article}

\usepackage[nonatbib,final]{dlde_neurips_2022}
\usepackage{graphicx}
\usepackage{caption}
\usepackage{subcaption}

\usepackage{amstext,amsthm,amssymb,amsmath} 
\usepackage{hyperref} 

\newcommand{\dataset}{\ensuremath{\mathcal{D}}}
\newcommand{\loss}{\ensuremath{\mathcal{L}}}
\newcommand{\ntrain}{\ensuremath{N}}
\newcommand{\R}{\ensuremath{\mathbb{R}}}
\newcommand{\x}{\ensuremath{\mathbf{x}}} 
\newcommand{\tx}{\ensuremath{\tilde{\mathbf{x}}}} 
\newcommand{\eps}{\ensuremath{\epsilon}}

\newcommand{\params}{\ensuremath{\boldsymbol{\theta}}}
\newcommand{\nnet}{\ensuremath{f_{\params}}}

\title{Experimental study of Neural ODE training with adaptive solver
  for dynamical systems modeling}

\author{%
  Alexandre Allauzen $^{1,2}$, Thiago Petrilli Maffei Dardis$^{2}$, Hannah Plath$^{2}$\\
  $^{1}$LAMSADE CNRS, Dauphine University-PSL, $^{2}$ESPCI-PSL \\
  \texttt{alexandre.allauzen@dauphine.psl.eu}
}

\begin{document}

\maketitle

\begin{abstract}
  Neural Ordinary Differential Equations (ODEs) was recently
  introduced as a new family of neural network models, which relies on
  black-box ODE solvers for inference and training. Some ODE solvers
  called adaptive can adapt their evaluation strategy depending on the
  complexity of the problem at hand, opening great perspectives in
  machine learning. However, this paper describes a simple set of
  experiments to show why adaptive solvers cannot be seamlessly
  leveraged as a black-box for dynamical systems modelling. By taking
  the Lorenz'63 system as a showcase, we show that a naive application
  of the Fehlberg's method does not yield the expected
  results. Moreover, a simple workaround is proposed that assumes a
  tighter interaction between the solver and the training strategy.
\end{abstract}

\section{Introduction}
\label{sec:intro}

A recent line of work has explored the interpretation of residual
neural networks~\cite{He16Deep} as a parameter estimation problem of
nonlinear dynamical
systems~\cite{Haber17Stable,WE17Proposal,Lu18Beyond}.  Reconsidering
this architecture as an Euler discretization of a continuous system
yields to the trend around Neural Ordinary Differential
Equation~\cite{Chen18NODE}.  This new perspective on deep learning
holds the promise to leverage the decades of research on numerical
methods. This can have many benefits such as parameter efficiency and
accurate time-series modeling, among others.

Numerical solvers for ODE act as a bridge between the continuous time
dynamical system and its discrete counterpart that builds the deep
neural network. As introduced by~\cite{Chen18NODE}, modern ODE solvers
can provide important guarantees about the approximation error while
adapting their step size used for integration. Therefore the cost of
evaluating a model scales with problem complexity, in theory at
least. This paper address this question empirically using the
Lorenz'63 system described in section~\ref{sec:node} as a testbed for
dynamical systems modelling. Our neural ODE framework relies on
Fehlberg's integration method which proposes a strategy of step-size
adaptation as summarized in~\ref {ssec:fehlberg}. A first round of
experiments is reported in~\ref{sec:first-exp} to assess how this
adaptive method impacts the inference of the model and the training
process. We show empirically that the adaptive strategy is in fact
ignored.  In section~\ref{sec:fehlberg-training}, a simple solution,
called \textit{Fehlberg's training}, is introduced that requires to
open the black-box of the solver for a tighter interaction with the
training process\footnote{The code to reproduce the
  results:\url{https://github.com/Allauzen/adaptive-step-size-neural-ode}}.
 
\section{Neural ODE applied to the Lorenz'63}
\label{sec:node}

To empirically analyse how an adaptive solver interacts with the
training procedure of a neural ODE model, we consider the
Lorenz'63 system~\cite{Lorenz63}. This ``butterfly'' attractor was
originally introduced to mimic the thermal convection in the
atmosphere, but nowadays this chaotic system is broadly used
as a benchmark for time series modeling. See for
instance~\cite{Nassar18Tree,Champion19Data,Greydanus19Hamlitonian,Dubois20Data,Gilpin21Chaos}
just to name a few recent work. Consider a point $\x \in \R^3$ with its three
coordinates $x_1, x_2, x_3$. The Lorenz'63 system consists of three
coupled nonlinear ODEs:
\begin{equation}
  \label{eq:lorenz}
  \dot{x}_1 = \frac{dx_1}{dt} = \sigma (x_2 - x_1),\phantom{5ex}
  \dot{x}_2 = x_1(\rho - x_3) - x_1,\phantom{5ex}
  \dot{x}_3 = x_1x_2 - \beta x_3.
\end{equation}
In this work we consider the standard setting
($\beta = 8/3,\ \sigma = 10,\ \rho = 28$), such that the solution
exhibits a chaotic regime. The dataset generation uses the explicit
Runge-Kutta (RK) method of order 5(4) with the Dormand-Prince
step-size adaptation in order to get an accurate integration.

\subsection{Neural ODE model}
\label{ssec:NODE-1}

The goal is to learn a generative model of this attractor, given a
training set $\dataset=(\tx_{i})_{i=1}^{\ntrain}$ made of $\ntrain$
examples. For time series modelling, recurrent
architectures~\cite{Nassar18Tree,Dubois20Data} or physically inspired
models~\cite{Greydanus19Hamlitonian} are often considered with
success. However, the system understudy derives from an ODE and Neural
ODE is also a well suited framework to consider.
The main idea is to learn the dynamics underlying the generation of
$\dataset$. The neural network aims at learning
$ \dot{\x} = \nnet (\x)$, where $\nnet$ is an arbitrary architecture
defined by its set of parameters $\params$. Inference with Neural ODE
thus requires a numerical solver denoted by $\textrm{ODE\_Solve}$ to
compute the output. In our case, we consider the prediction task of
the point ${\x}_{i}$ at time $i$, given the previous training point
$\tx_{i-1}$, such that ${\x}_{i}=\textrm{ODE\_Solve}(\nnet, \tx_i)$.
The model is learnt by minimizing the Mean-Squared-Error:
$\loss(\params, \dataset) = \sum_{i=1}^{N} ||\tilde{\x}_i - \x_i
||^2$.

An advantage of Neural ODE is the  choice of the solver and
especially the possibility to adapt the step size depending on the
problem complexity. In this paper, we focus on the Fehlberg's 3(2)\footnote{The Runge-Kutta method of order 3 (RK3)
  along with second order version for the error control.}
method~\cite{Fehlberg68}, for its
simplicity of exposition, since our goal is to analyse  how the interaction between the solver
and the training process.

\subsection{Fehlberg's method under the hood}
\label{ssec:fehlberg}

In general, variable step size methods all rely on the same idea: for
an inference step, try two different algorithms, giving two different
hypotheses called $A_1$ and $A_2$. The two algorithms are chosen so that:
i) the difference $A_2-A_1$ provides an approximate of the local
truncation error; and ii) both algorithms use at most the same
evaluations of $\nnet$ to limit the computational cost.  In our case,
the method requires three evaluations \footnote{In
  general, evaluations $f_1,f_2f_3$ are respectively associated to the
  time $t_i,t_{i+h},t_{i+h/2}$. In our case, the function $\nnet$ is
  time invariant and $t_i=i$. } of $\nnet$ defined as follows for the step size denoted by $h$:
\begin{equation*}
f_1 = \nnet(\x_i),\phantom{5ex} 
f_2 = \nnet(\x_i+hf_1), \phantom{5ex} 
f_3 = \nnet(\x_i+\frac{h}{4}[f_1+f_2])
\end{equation*}
With these three evaluations, two approximations for the next point can be derived: 
\begin{equation}
  \label{eq:hyps}
   A_1 = \x_i+\frac{h}{2}[f_1+f_2] \text{ (RK2 method), and } A_2 = \x_i+\frac{h}{6}[f_1+f_2+4f_3] \text{ (RK3).} 
\end{equation}
Considering the property of these two algorithms, we can estimate the
following error rate:
\begin{equation}
  \label{eq:error}
  r = \frac{|A_1-A_2|}{h} \approx Kh^2
\end{equation}
If this error rate exceeds a chosen tolerance $\eps$, the hypothesis
$A_2$ is rejected and we need to restart the computation with a new
step size: $ h'= S \times h\sqrt{\epsilon/r}, $ with $S$ a safety
factor. The initial time step is $h=1$, and we use the default value:
$S=0.9$ and $\eps=0.1$. This strategy allows the model to increase the number
of integration steps\footnote{In practice, $1/h'$ is rounded up to
  determine the number of steps.} when predicting the next point
$\x_{i+1}$ to adapt the expected precision given $r$.

\section{First round of experiments}
\label{sec:first-exp}

\begin{figure}[thb]
  \hspace*{-2cm}
  \includegraphics[width=1.2\textwidth]{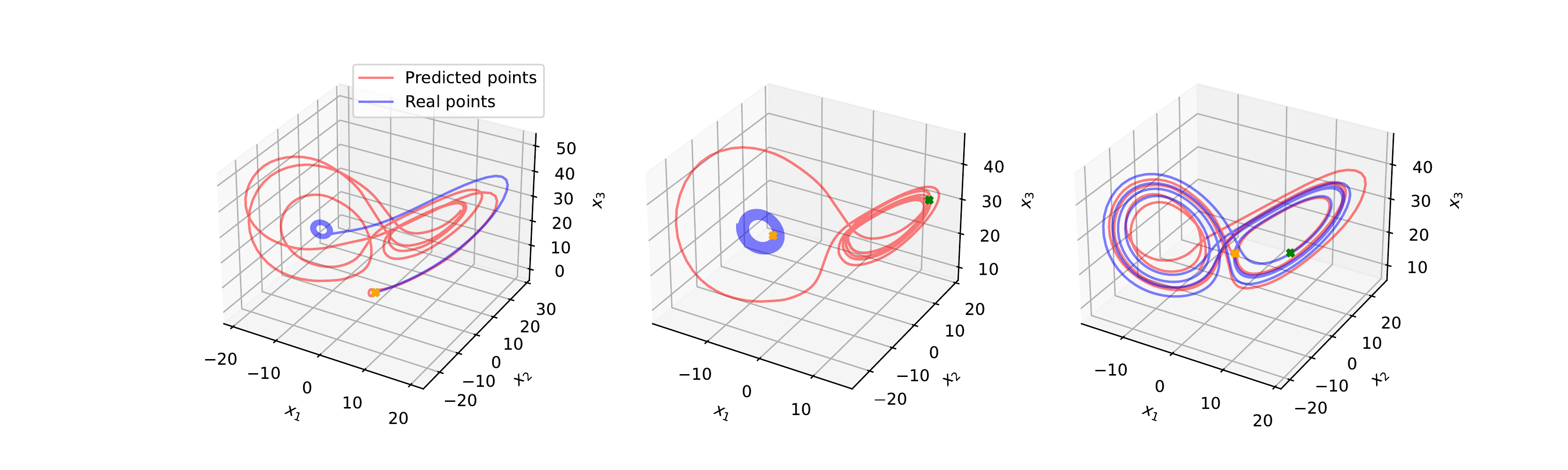}
  \caption{The trajectories predicted after a regular training, using
    the Fehlberg's solver as a black box. Each figure depicts a
    different time slice of the generated trajectory and of the
    original training data: from 0 to 600, 600 to 1200 and 2000 to 2600. }
  \label{fig:baseline_lorenz}
\end{figure}

For the first set of experiments, a simple ODE model is trained with
L-BFGS~\cite{Liu89LBFGS} wrapped in the Fehlberg's method. The neural
network $\nnet$ is a simple feed-forward architecture with two hidden
layer of size $50$ and ReLu activation. The trained model is then used
to generate data, starting from the same initial condition and the
figure~\ref{fig:baseline_lorenz} shows the results obtained for
different time windows. This task differs from the training phase, and
we can see that the model fails to accurately reproduce the trajectory
of the original dynamical system. Especially the beginning of the
trajectory greatly differs. The combination of the accumulated
truncation errors with the chaoticity of the Lorenz'63 leads to too
challenging generation task while the model can capture some important
features like the ``butterfly'' aspect. The same trend is observed on
the development set.

\begin{figure}[thb]
  \centering
  \includegraphics[width=1\textwidth]{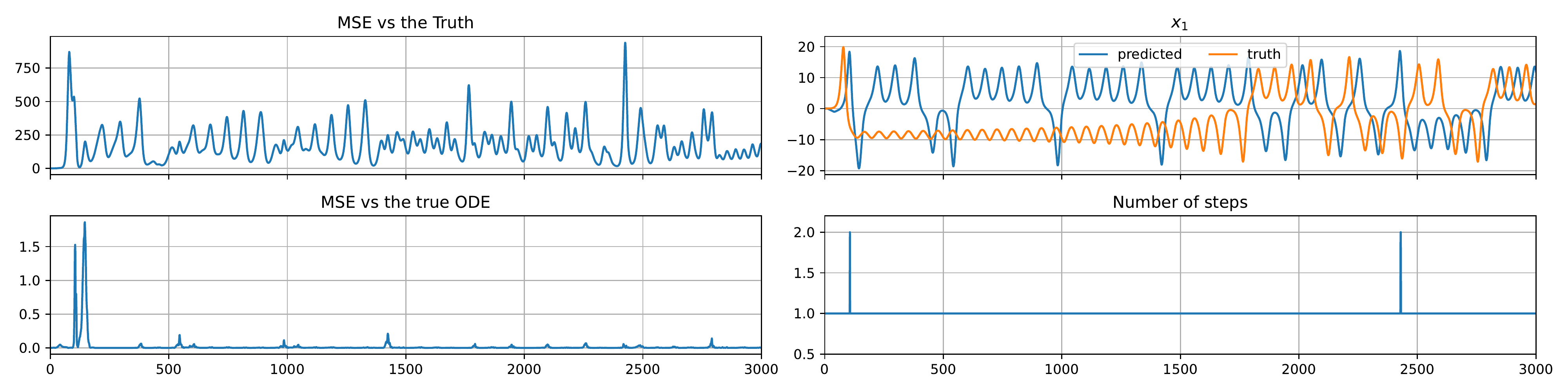}
  \caption{Time evolution of (from left to right and top to bottom):
    the MSE, the evolution of $x_1$, the MSE \textit{w.r.t} the true ODE of
    Lorenz'63, and  the number of
    steps.}
  \label{fig:baseline_series}
\end{figure}

This disappointing performance can be explain by the
figure~\ref{fig:baseline_series} which represents the time evolution
of different quantities of interest. The evolution of the MSE
highlights its limitation to evaluate chaotic system modelling:
errors on the first points are small in terms of MSE, while 
inducing an important time distortion later. This can be observed by comparing the
trajectories of the first component $x_1$. Since we have access to the
physical system, we can measure another MSE by comparing each point
$\x_i$ generated by the model, with what would be generated by the
Lorenz ODE of equation~\ref{eq:lorenz} from the previous predicted
point or $\x_i^*=Lorenz(\x_{i-1})$. This quantity is the third time
series represented in figure~\ref{fig:baseline_series} and provides a
different insight on the performance.  More importantly, the last plot
monitors the number of integration steps for each prediction. In most
of the case, this number is stuck to 1, showing a very limited usage
of the step size adaption.

To further understand this fact, we investigate what happens during
the training process by monitoring three different quantities measured
after each epoch and represented in figure~\ref{fig:training_series}. 
The first column corresponds to the standard training method which
considers the adaptive solver as a black-box as described in the
seminal paper~\cite{Chen18NODE}. We can observe from the second row
that the Fehlberg's method accepts the vast majority of the RK3
prediction with one step, without resorting to an adaptive step
size. Moreover, the step size is just slightly increased for a handful
of training examples (see the third row). We can conclude that the
adaptive step size is not harnessed by the Neural ODE model.

In fact, the model is randomly initialized with values around zeros,
and the error estimation in equation~\ref{eq:error} falls under $\eps$
in most of the case, and whatever the input of the model is. A first
workaround would be to explore a tailored initialization scheme, with
the goal of increasing this initial estimate. Another workaround
consists in lowering the tolerance factor $\eps$ to let the Felhberg's
method reject more hypotheses and therefore adapt the step size more
frequently.  However, this introduces a new hyperparameter to
tune\footnote{Additional experiments in Appendix~\ref{sec:eps} show the limits of this remedy.}. 

\section{Fehlberg's training }
\label{sec:fehlberg-training}

\begin{figure}[thb]
  \centering
  \includegraphics[width=0.49\textwidth]{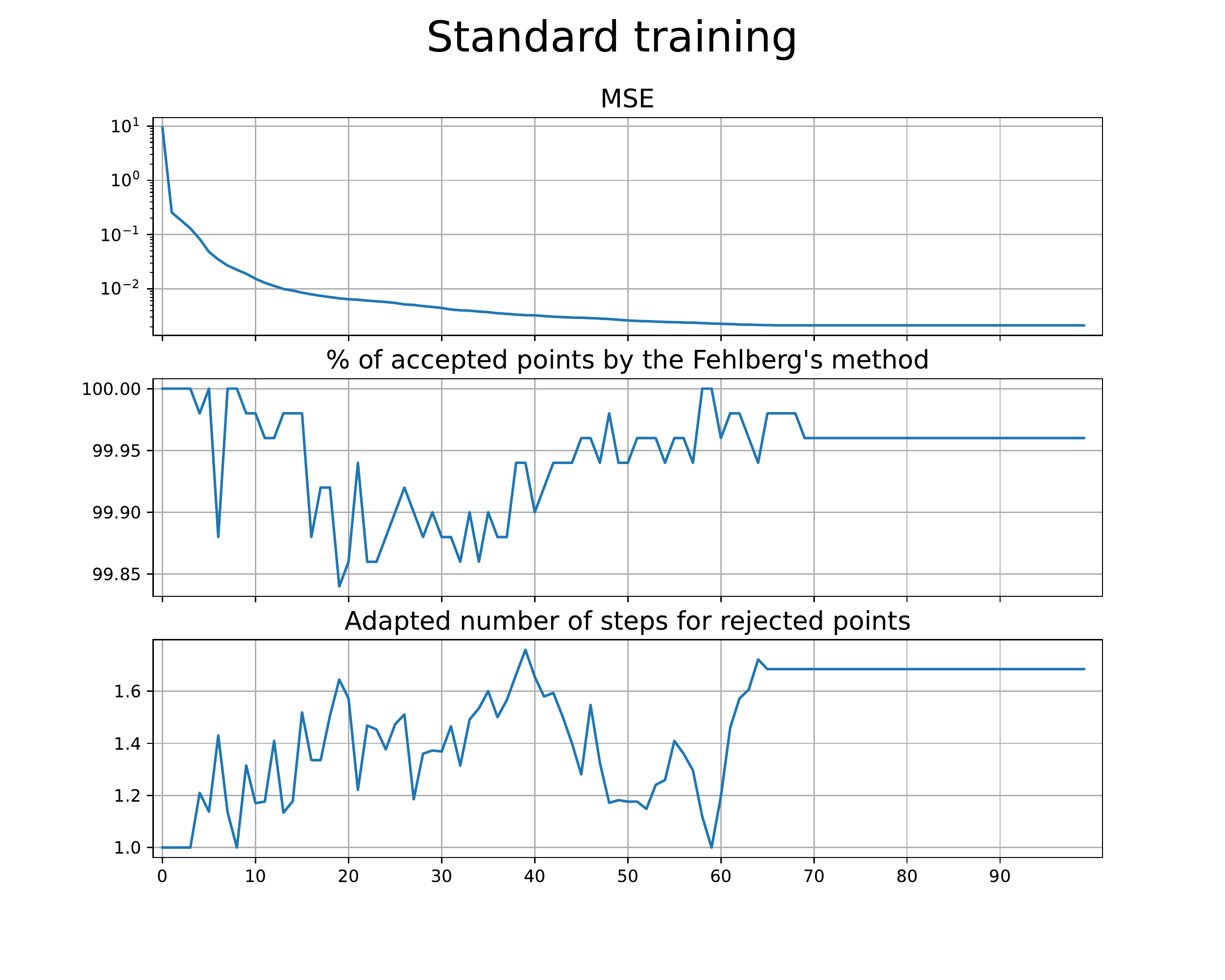}
  \includegraphics[width=0.49\textwidth]{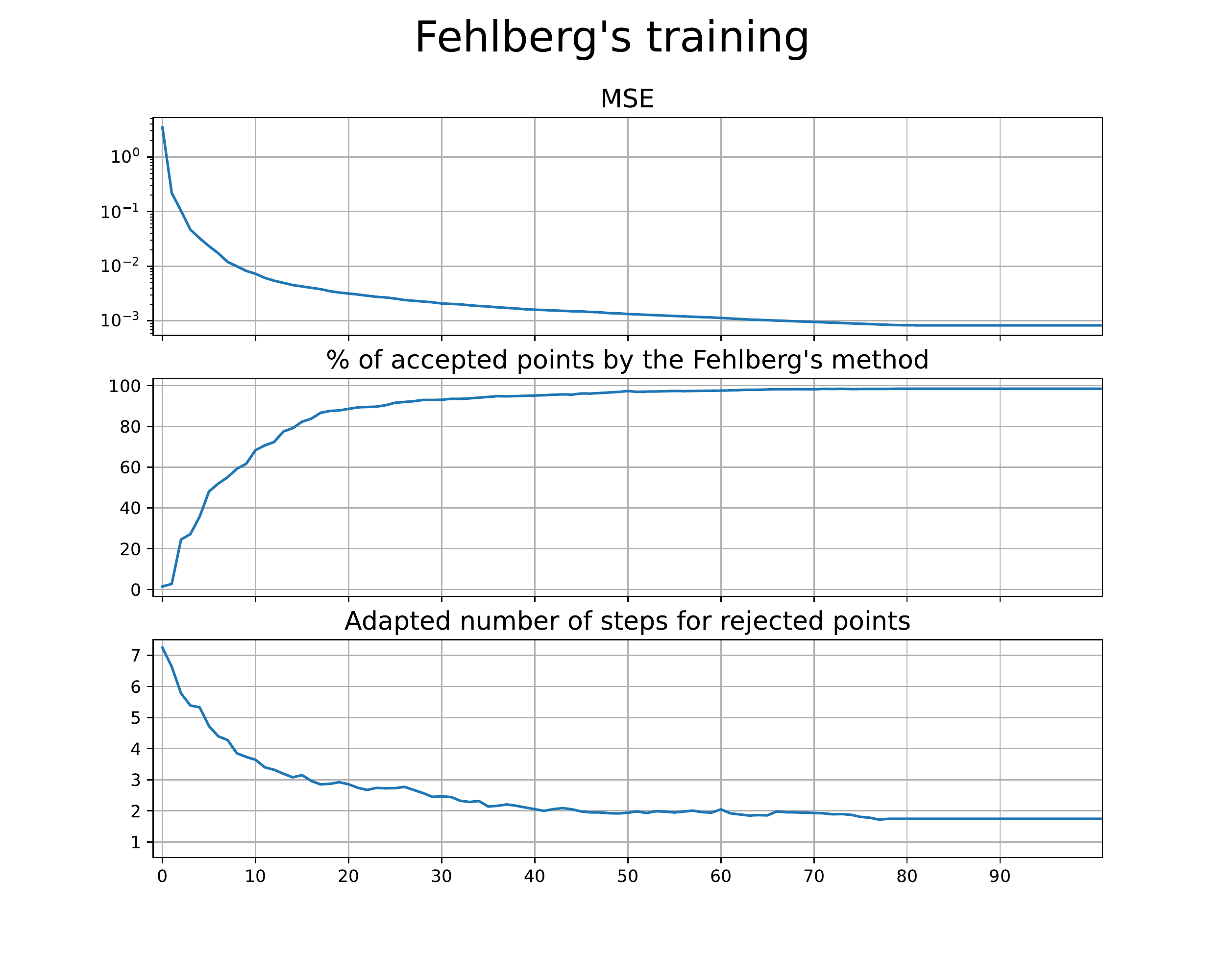}
  \caption{Time evolution for two training conditions of: the MSE loss; the percentage of
    accepted hypotheses $A_2$; the new number of steps for the rejected
    hypotheses (before rounding).}
  \label{fig:training_series}
\end{figure}

In this paper, we propose another solution which modifies the training
procedure. The idea derives from the Fehlberg's method by simply
changing how the local truncation error is estimated. Let us consider equation~\ref{eq:hyps} defining the error rate.
When predicting the value $\x_i$ from $\tx_{i-1}$, we can use directly the target value $\tx_i$, which is available during training,  instead of the raw estimate $A_1$ given by the Heun's method (\textit{aka} RK2). The basic
hypothesis $A_2$ is still obtained with one step of RK3.
Given this modified error rate, the step-size adaptation remains
unchanged. This new method greatly impacts the training process. On
the second column of the figure~\ref{fig:training_series}, we observe
a very different trend: the proportion of accepted hypotheses starts at 0 and 
progressively increases to reach approximately 98\%.
For the rejected hypotheses, the new number of steps (before rounding)
starts at the high value of $7$ and affects all the training examples:
after the random initialization of $\nnet$, all the training examples
are considered as difficult, while the high number of steps multiplies
the amount of updates for $\params$.  Then the number of steps
smoothly decreases to converge to a value just under $2$ and affects
about 2\% of the training points. We can conclude that the new
training scheme allows the Neural ODE model to really leverage the
adaptive step size strategy. Of course it requires to interact with
the adaptive solver, but without adding new hyperparameters or
trade-offs. See Appendix~\ref{sec:add_fehlberg} for more figures and
comparisons.

\section{Conclusion}
In this experimental paper, we investigated how the numerical solver
interacts with the training process of a neural ODE model. We focused
our work on a solver able to adapt its step size.  With a simple
experimental setup, the results showed that using a solver as black
box drastically hinders the promise of the adaptive strategy for the
step size. This is a real issue to model more complex dynamical
systems. We proposed a simple yet efficient solution that requires a
tighter interaction between the solver and the training process. With
our results, it will be possible to tackle more challenging tasks and
while we focused on generative models for time series forecasting, it
could be useful to extend our approach to classification tasks.

\textbf{Aknowledgment:} This work was funded by the French National Research Agency (ANR
SPEED-20-CE23-0025). Thanks to Clément Royer for his help.


\bibliographystyle{plain}
\bibliography{aa_dlde22}

\clearpage
\appendix

\section{Experimental setup}
\label{sec:setup}
For the experiments reported in the paper, two datasets were generated, one for training and one for testing. Each
of them consists in a trajectory of 5000 points.

\subsection{Optimization}
\label{ssec:optim}

The training process is carried out in batch mode with  the L-BFGS optimizer with  the default setting: 
\begin{itemize}
\item The initial learning rate is set by default to 1. However,
  L-BFGS quickly reconsider this value through many function
  evaluation.
\item The number of iterations per optimization step is set to 20. It
  means that one epoch of the batch training is not comparable with
  other optimization algorithms like gradient descent or ADAM.
\item   The number  of function evaluations per optimization step is 25, and the tolerance factors are set to $10^{-5}$ for the gradient, and  $10^{-9}$ for the tolerance on function value changes. 
\item The history size is limited to 100 and the optimizer uses a line search (see~\cite{Liu89LBFGS}).
\end{itemize}

For the Fehlberg's method, we use our own implementation without using
the adjoint method.

\subsection{Batch training}
\label{ssec:batch}

With Neural ODE, $\nnet$ represents the elementary block. The inference
step consists building on the fly the whole network, depending on how
the solver proceeds to predict the output. For instance, with the
Euler method, the whole network is very similar to the ResNet
architecture~\cite{He16Deep,Haber17Stable,Lu18Beyond}.  With the
Fehlberg's method, the step size is adapted for each training example,
meaning that the whole network (and its computational graph) is
different. This is an obstacle for mini-batch or batch training.
However, online training drastically increase the computational time.
As a trade-off, we propose the following procedure for each (mini)
batch:
\begin{itemize}
\item compute $A_1, A_2$, and $r$;
\item for the  ``accepted'' subpart, such as $r<\eps$,  return $A_2$;
\item for the remaining part, compute the new step size $h'$ for each
  training example and keep the minimum value, clipping the value at
  $1/10$, which corresponds to $10$ steps of integration. Recompute
  $A_2$ with this value.
\end{itemize}
With this procedure, we can therefore benefit from batch training and
the choice of the minimum value is a way to promote small step size.

\section{Impact of $\eps$}
\label{sec:eps}
\begin{figure}[thb]
  \centering
  \includegraphics[width=0.49\textwidth]{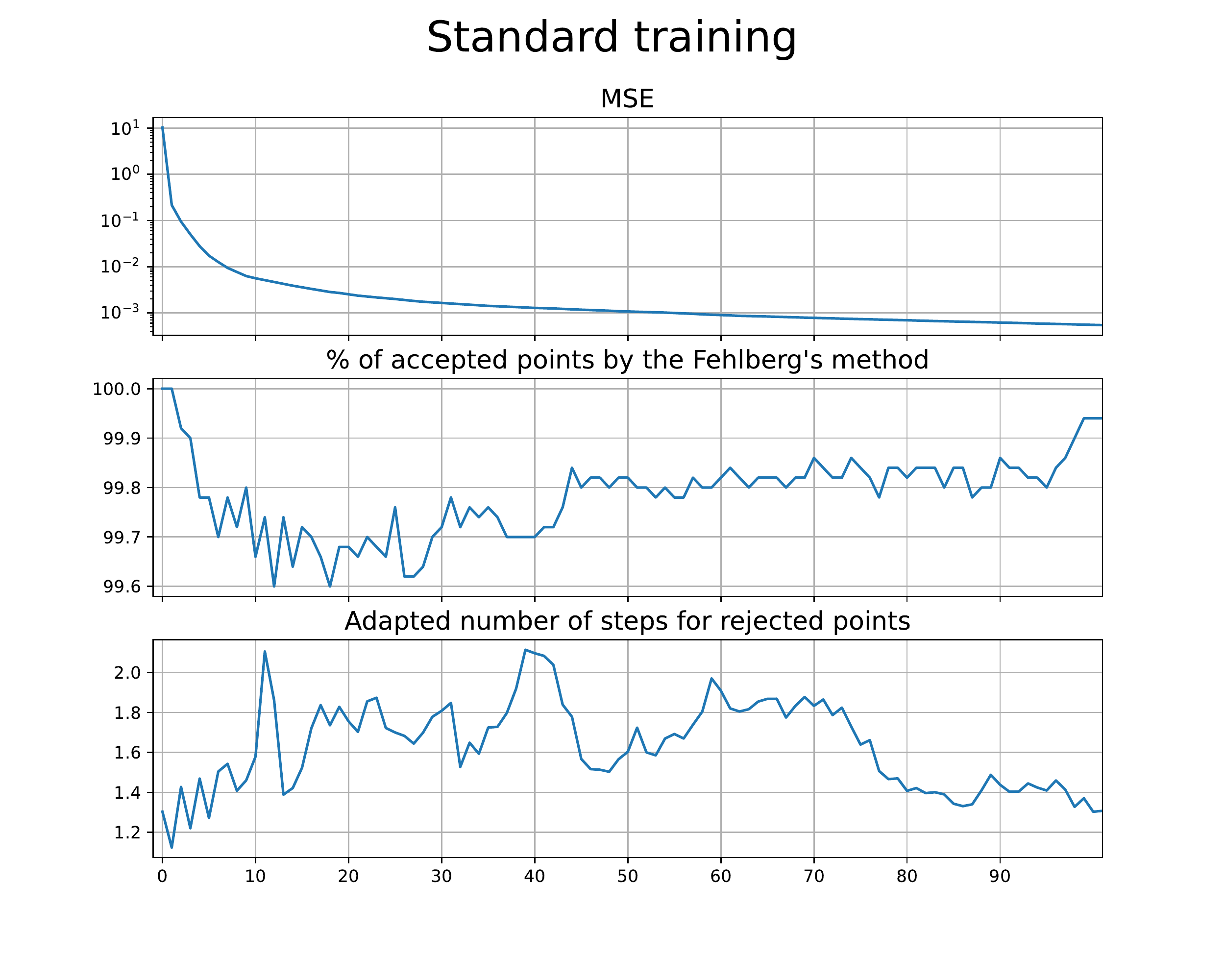}
  \includegraphics[width=0.49\textwidth]{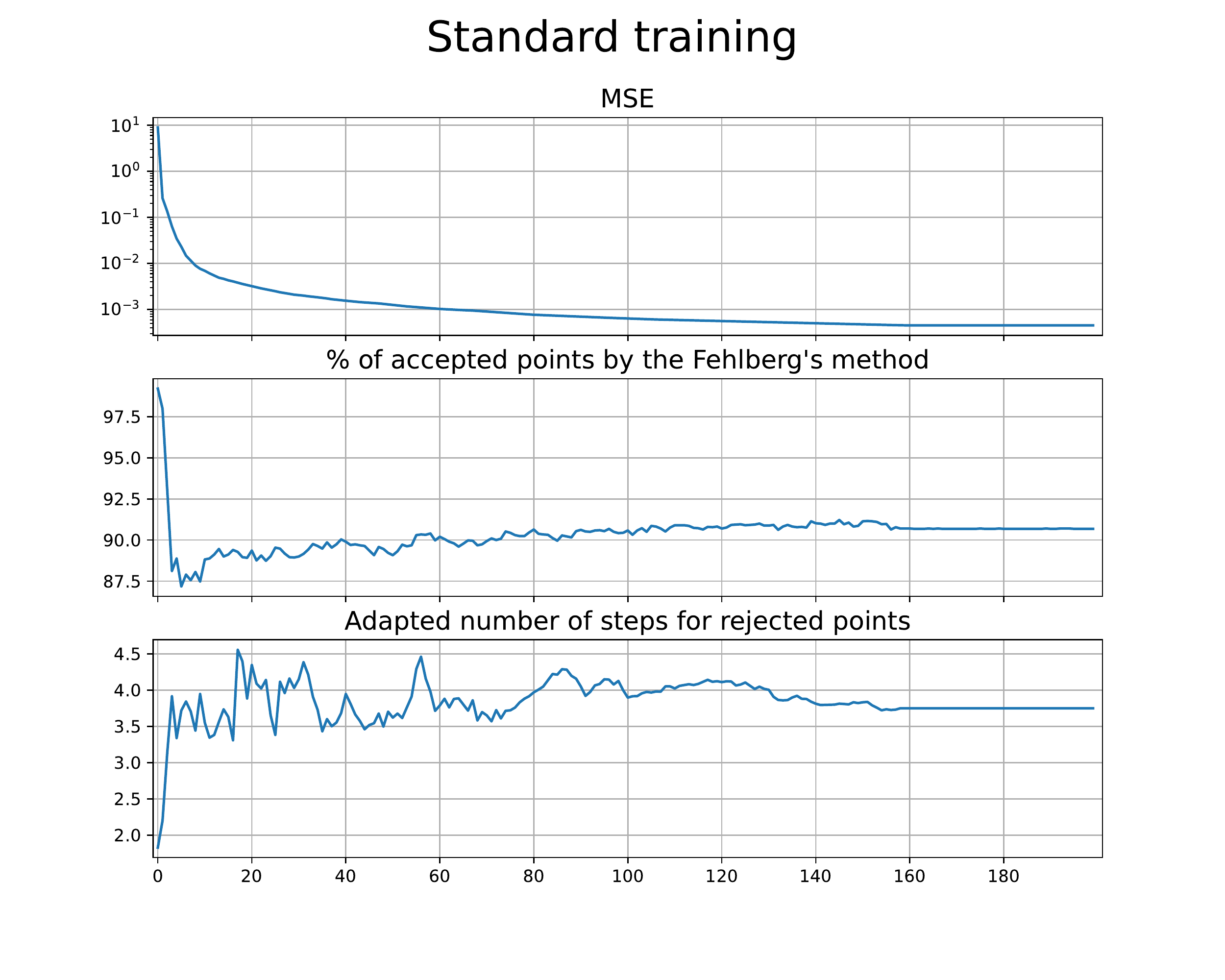}
  \caption{Time evolution for regular training with $\eps=0.05$ on the
    left, and $\eps=0.01$ on the right, for: the MSE loss; the
    percentage of accepted hypotheses $A_2$; the new number of steps
    for the rejected hypotheses (before rounding).}
  \label{fig:training_series_eps}
\end{figure}

The threshold $\eps$ introduced in section~\ref{sec:first-exp} can
mitigate the number of accepted and rejected hypotheses during the
Fehlberg's integration. To increase the number of rejected hypotheses,
we can try to lower $\eps$. To see how this hyperparameter can impact
the learning process, we train the same model with different values of
$\eps$. In the main part of the paper, the first column of
figure~\ref{fig:training_series} shows how the regular training
evolves with $\eps=0.1$ (the default value). Let us consider lower
values starting with $\eps=0.05$ in
figure~\ref{fig:training_series_eps}. In this case, it does not really
promote the step size adaptation. If we lower $\eps$ to $0.01$ as
shown on the right side of figure~\ref{fig:training_series_eps}, the
number of accepted hypotheses per epoch decrease around $90\%$, while
the number of steps is rounded up $4$ for the rejected hypotheses.

\begin{figure}[thb]
  \hspace*{-2cm}
  \includegraphics[width=1.2\textwidth]{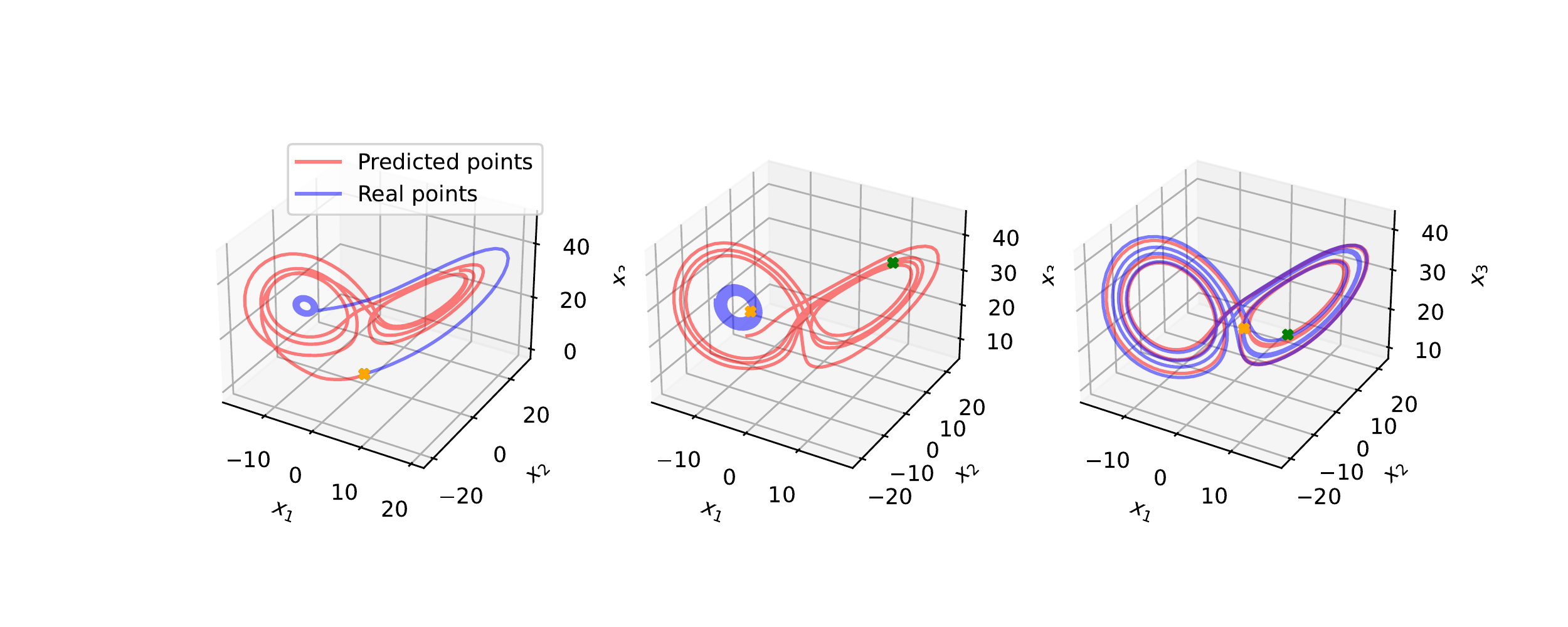}
  \caption{The trajectories predicted after a regular training with $\eps=0.01$, using
    the Fehlberg's solver as a black box. Each figure depicts a
    different time slice of the generated trajectory and of the
    original training data: from 0 to 600, 600 to 1200 and 2000 to 2600. }
  \label{fig:baseline_lorenz_eps}
\end{figure}

It is worth noticing that the $x$-axis is twice wider. The training
process is longer, in terms of number of epochs, and the computation
time is also augmented since $90\%$ of the training points requires 4
integration steps. However, if we look at the generation abilities,
the results are really better as shown in
figure~\ref{fig:baseline_lorenz_eps}: the generation still fails to
mimic the original Lorenz for the starting points, but after the
generation is really similar to the original one.

\section{Additional figures for Fehlberg's training}
\label{sec:add_fehlberg}
To complement the comparison of the baseline training and the
Fehlberg's training (see respectively section~\ref{sec:first-exp} and
\ref{sec:fehlberg-training}), additional figures are provided here.
\begin{figure}[thb]
  \includegraphics[width=1\textwidth]{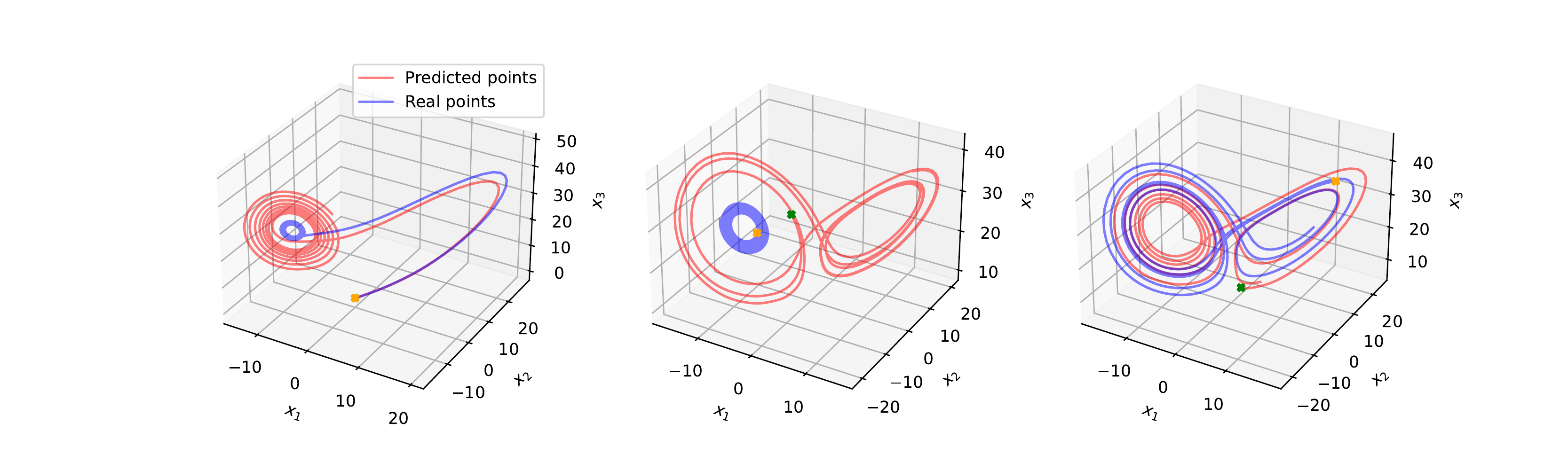}
  \caption{The trajectories predicted after the Fehlberg's training.}
  \label{fig:fehlberg_lorenz}
\end{figure}
\begin{figure}[thb]
  \centering
  \includegraphics[width=0.45\textwidth]{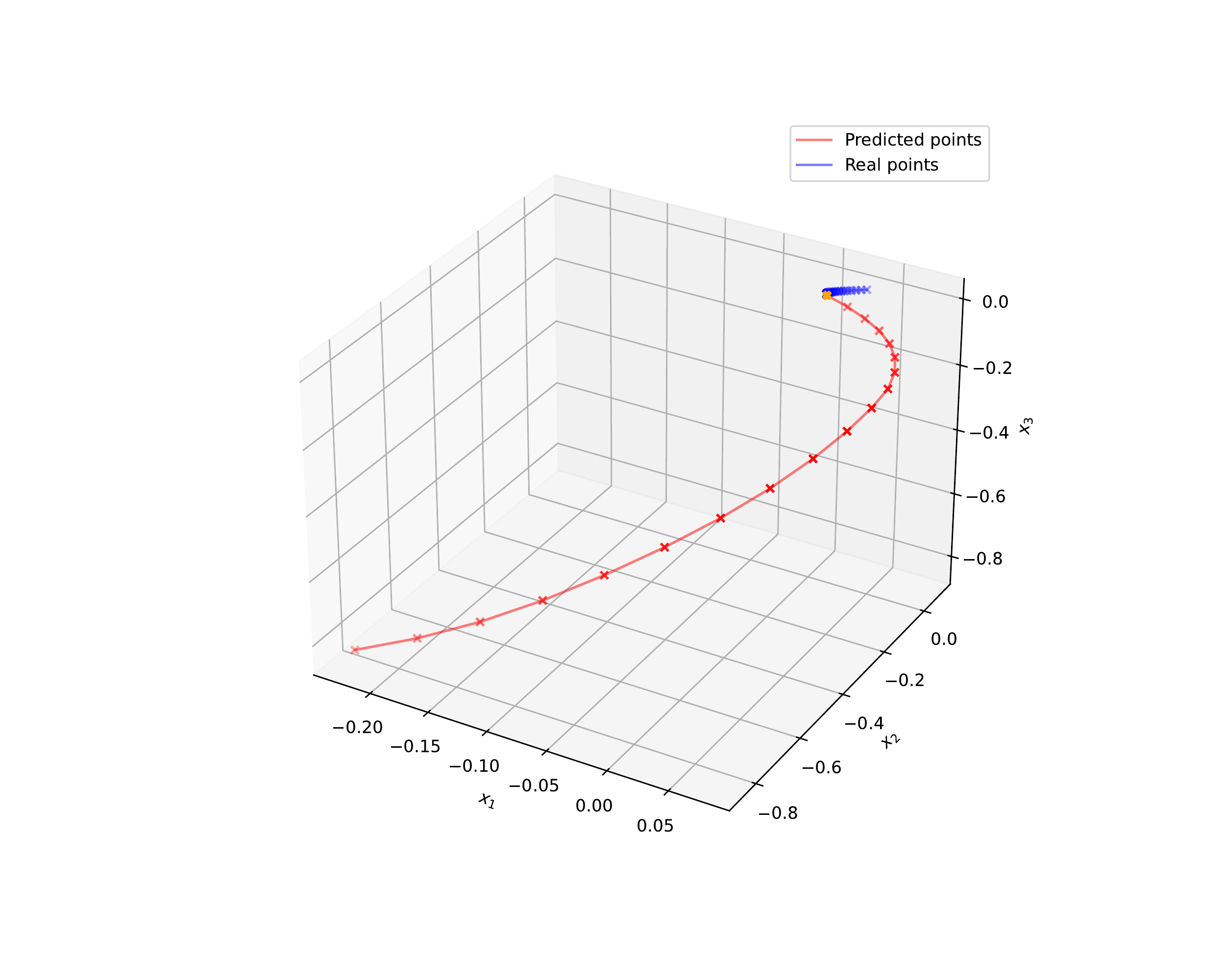}
  \includegraphics[width=0.45\textwidth]{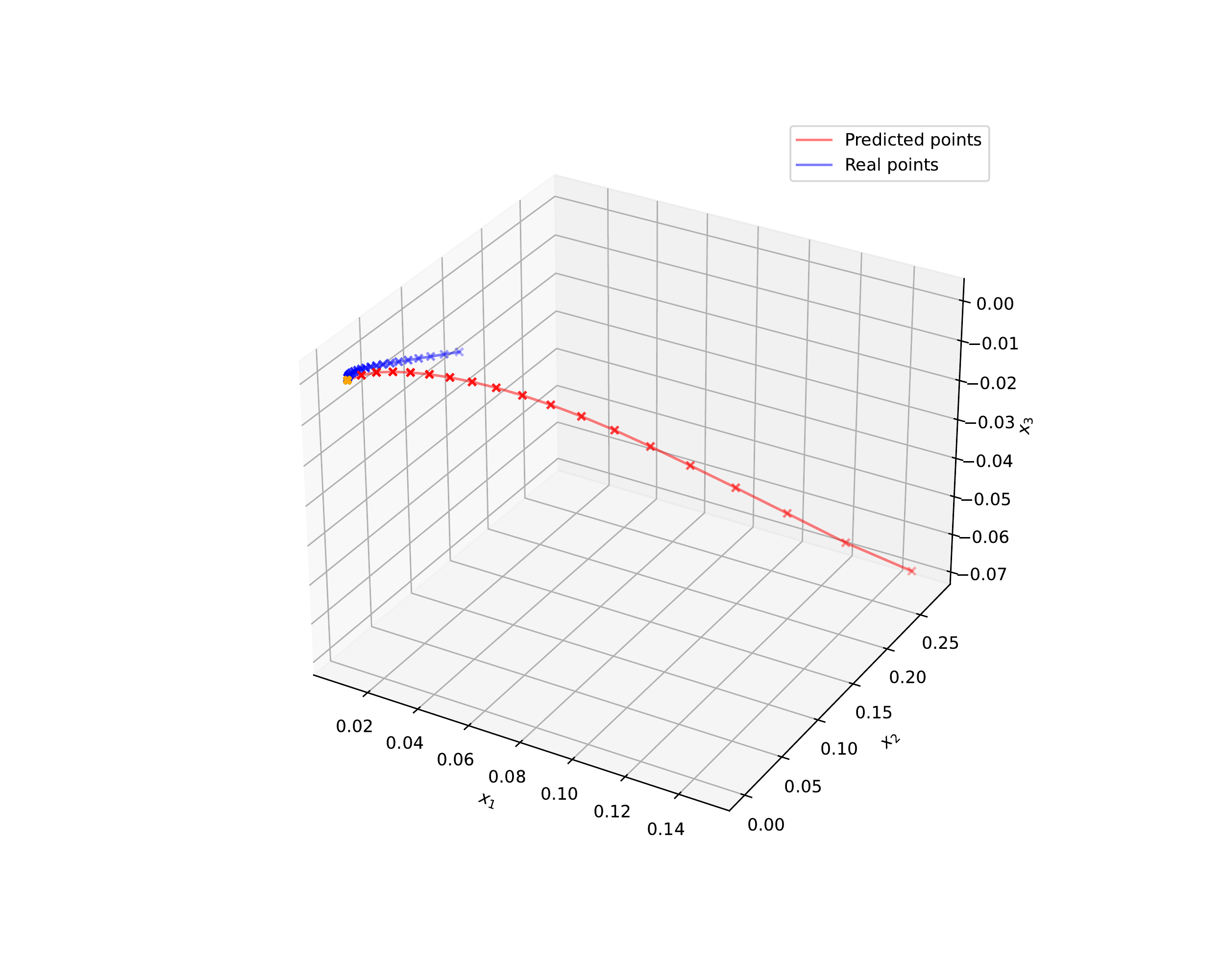}\\
  \includegraphics[width=0.45\textwidth]{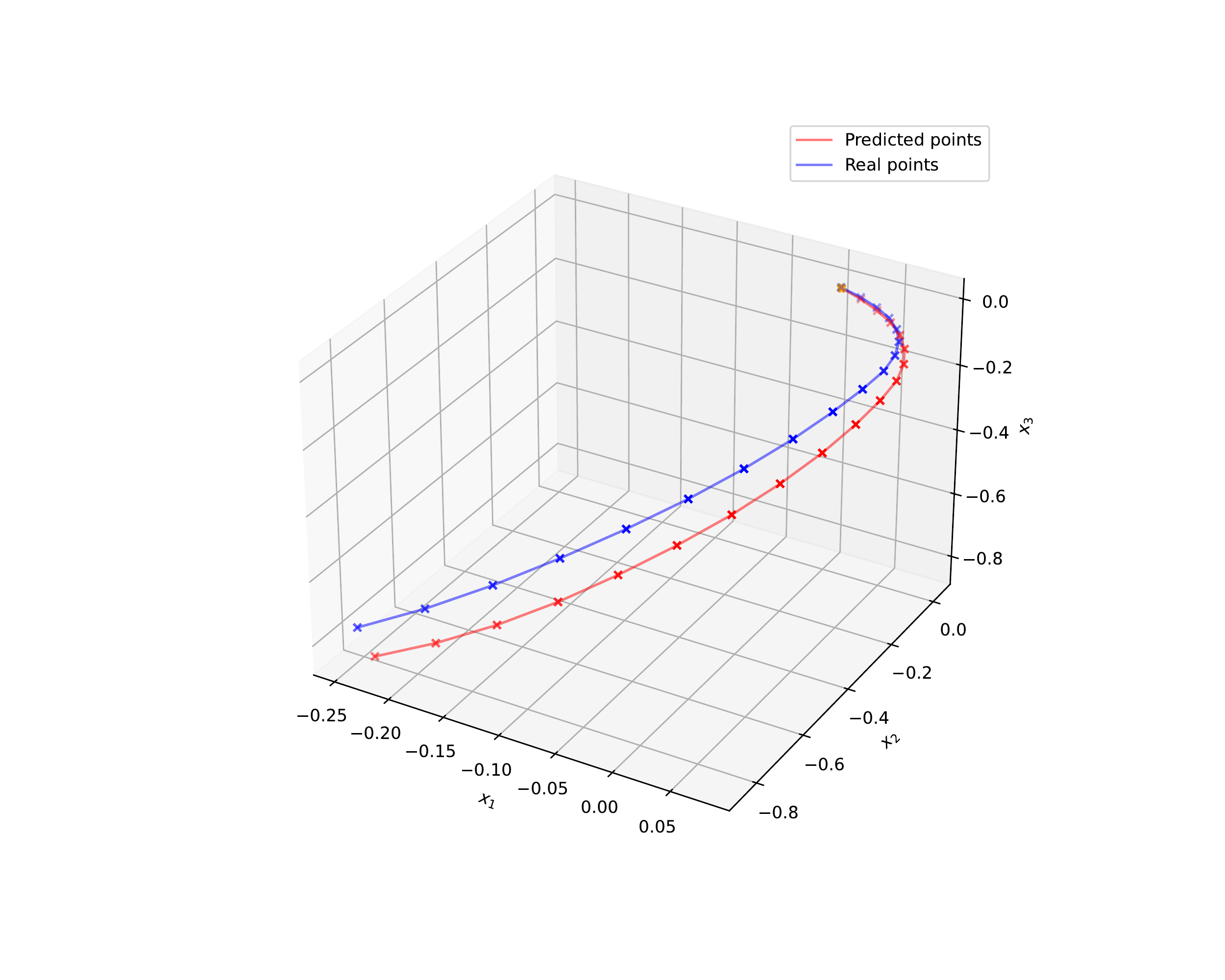}
  \includegraphics[width=0.45\textwidth]{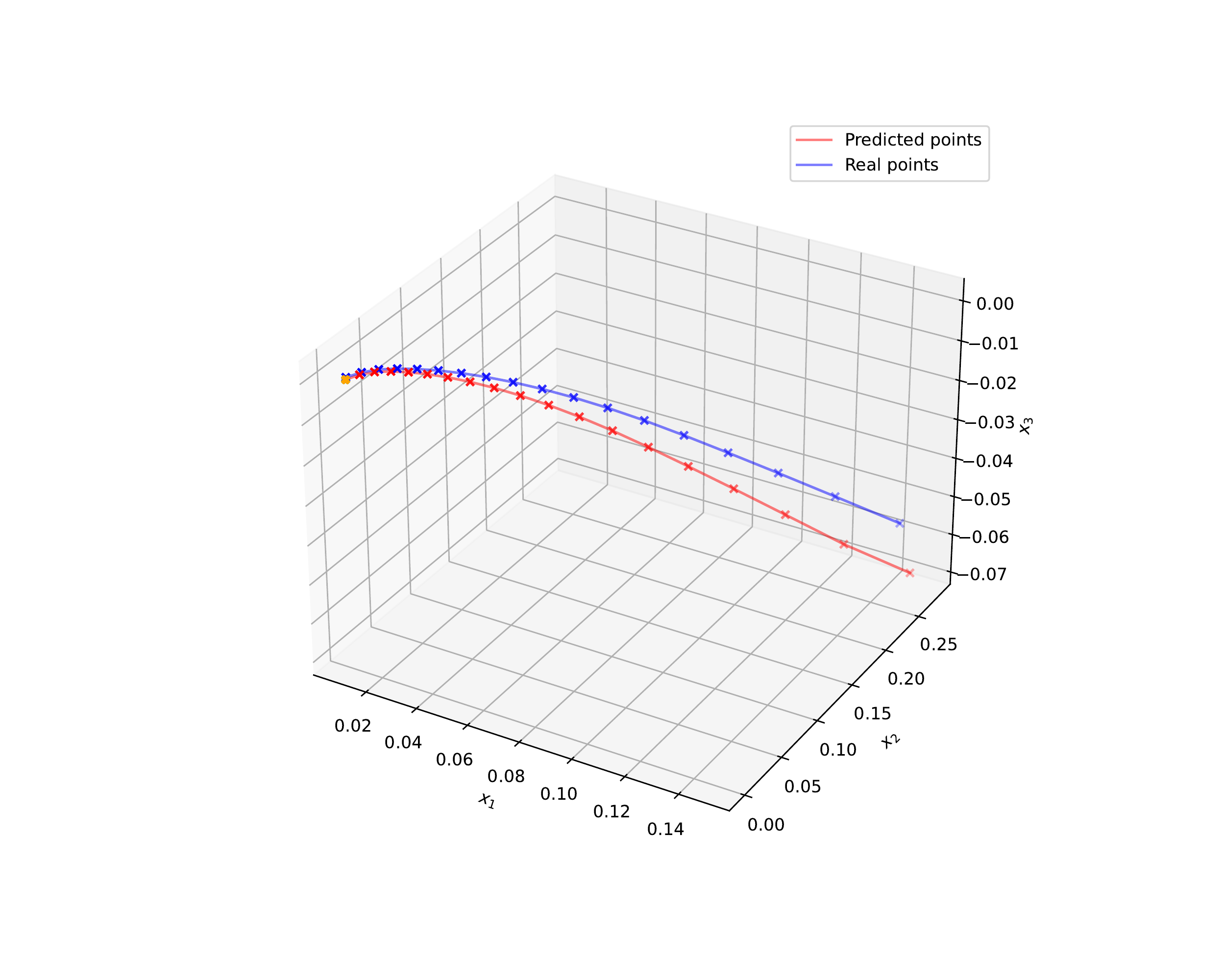}\\
  \includegraphics[width=0.45\textwidth]{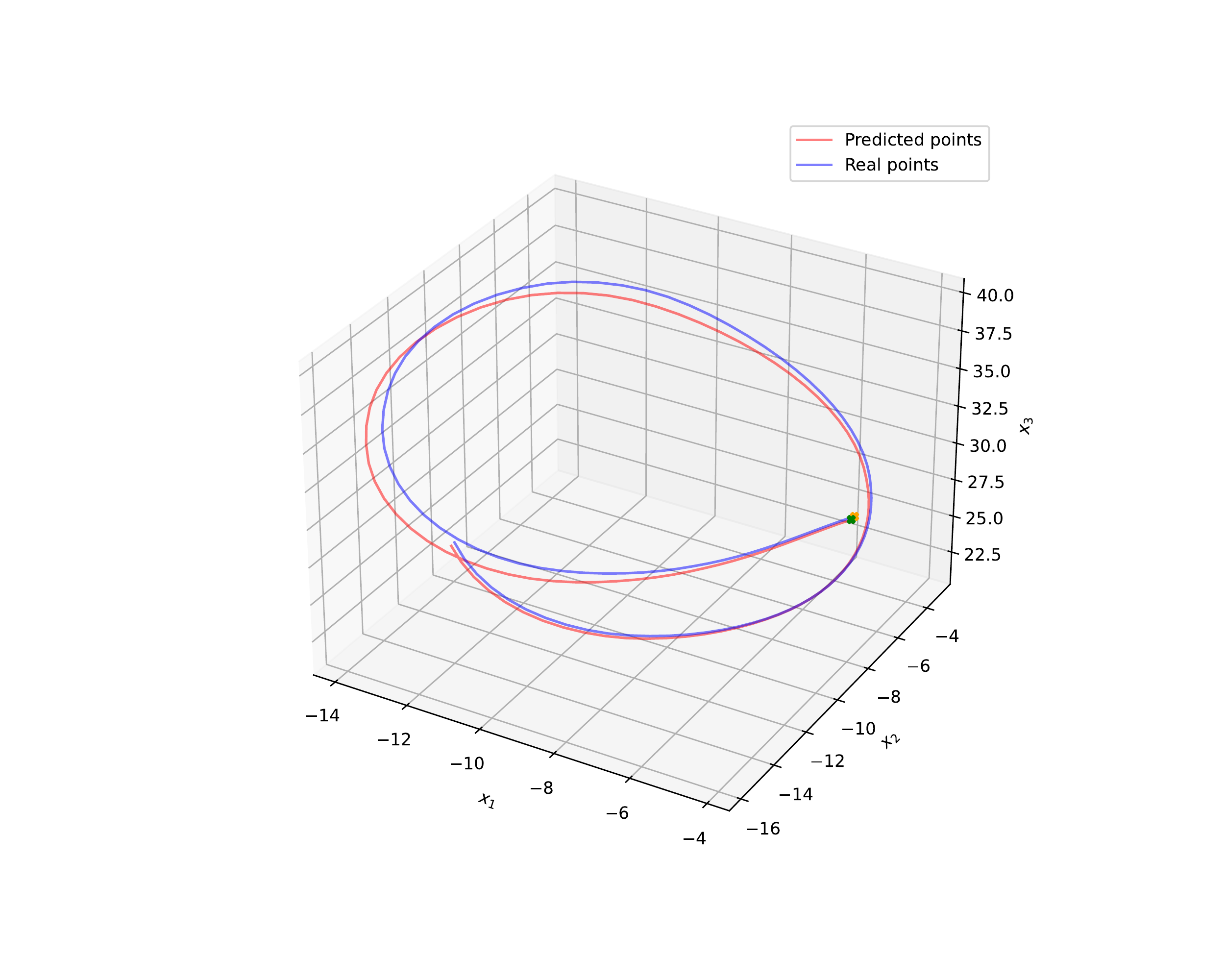}
  \includegraphics[width=0.45\textwidth]{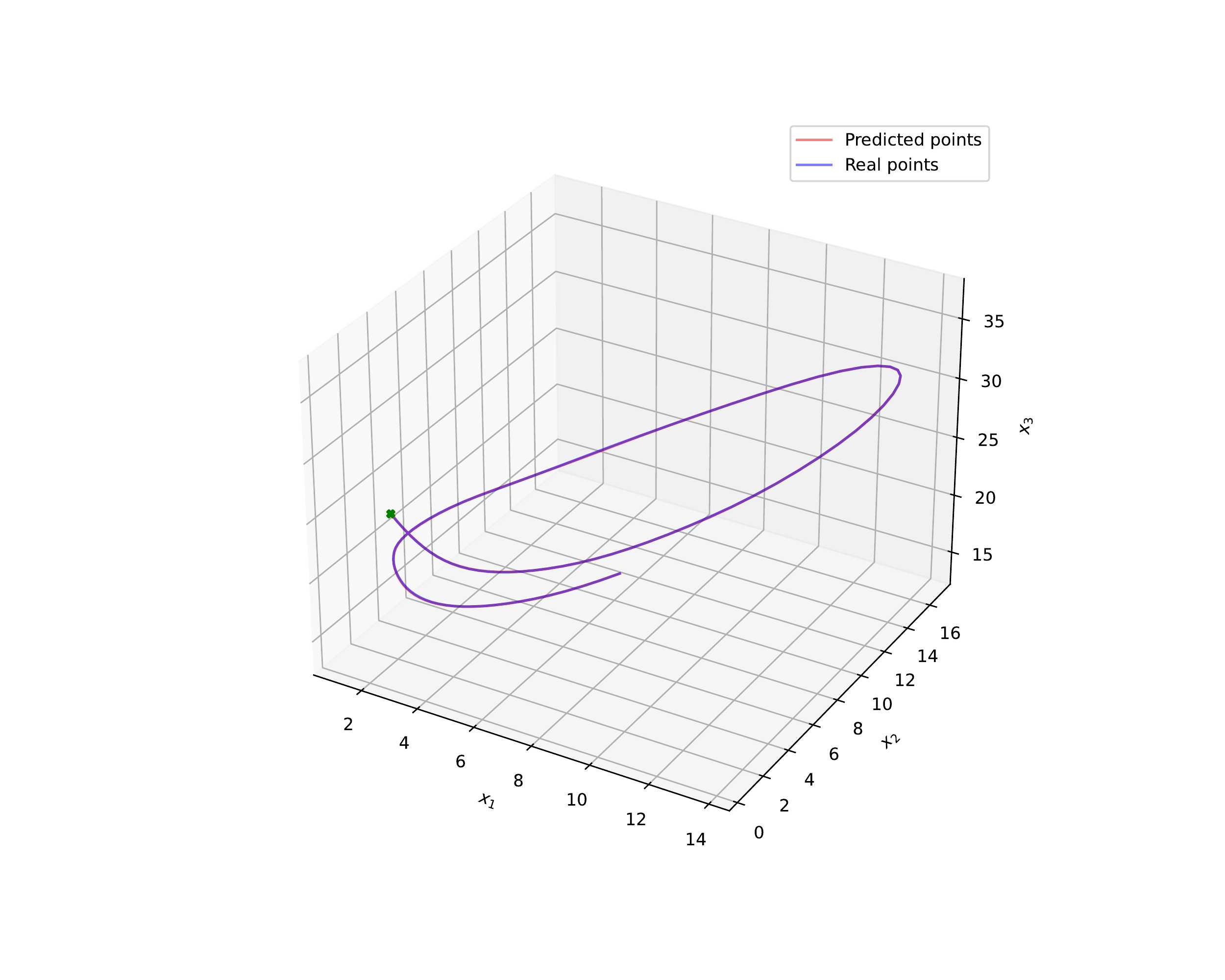}\\
  \caption{On the first row: the 20 first time steps of generation for
    the baseline and Fehlberg's training \textit{vs} the real data
    points. Then on the second row, the same predicted points
    \textit{vs} the points generated by the Lorenz ODE from the
    predicted ones (\textit{i.e} the ones used to estimate the
    modified MSE as described in section~\ref{sec:first-exp}). These
    two figures show the errors made by the models but these errors
    are not affected by the chaoticity of the Lorenz. The last row
    shows the same two figures, but for time steps from 1000 to 1100.}
  \label{fig:generations}
\end{figure}
\end{document}